# Assessing GPTZero's Accuracy in Identifying AI vs. Human-Written Essays

Selin Dik  1selindik@gmail.com; Osman Erdem  oerdem995@gmail.com; and Mehmet Dik
mdik@rockford.edu

## 1. Abstract

As the use of AI tools by students has become more prevalent, instructors have started using AI detection tools like GPTZero and QuillBot to detect AI written text. However, the reliability of these detectors remains uncertain. In our study, we focused mostly on the success rate of GPTZero, the most-used AI detector, in identifying AI-generated texts based on different lengths of randomly submitted essays: short (40-100 word count), medium (100-350 word count), and long (350-800 word count). We gathered a data set consisting of twenty-eight AI-generated papers and fifty human-written papers. With this randomized essay data, papers were individually plugged into GPTZero and measured for percentage of AI generation and confidence. A vast majority of the AI-generated papers were detected accurately (ranging from 91-100% AI believed generation), while the human generated essays fluctuated; there were a handful of false positives. These findings suggest that although GPTZero is effective at detecting purely AI-generated content, its reliability in distinguishing human-authored texts is limited. Educators should therefore exercise caution when relying solely on AI detection tools.

## 2. Introduction

Since the rise of generative AI, more and more students are using it as a means to get around work. AI detectors are now used to prevent misuse, and while they are generally effective, they occasionally produce false positives. The purpose of this experiment was to determine whether there was a correlation between false positives from AI detectors and the length of the text inputted into the system. In this experiment, we used GPTZero, made by the makers of ChatGPT. We tested this hypothesis using two groups of papers: AI generated, as our control group, and human written essays, as our experimental group.

We wanted to answer whether the length of the text inputted into GPTZero affected the likelihood of it being flagged as AI written. We hypothesized that as the length of text increased, the likelihood that the text getting flagged incorrectly would decrease. The results from this study could be used to assess the accuracy of AI detectors.  Therefore, we can ensure that students who are using AI in their work can be properly identified, and the students who are not, are not wrongly punished for using it.

## 3. Methodology

First, we divided the papers into two categories based on who they were written by: AI vs. human written. For human essays, we decided to use fifty student written essays that were written without the help of AI. For the AI essays, they were generated using random prompts using ChatGPT.

In total, we had fifty human written essays and twenty-eight AI written essays. We then sorted these essays into categories based on their lengths. We grouped the essays that were shorter than one hundred words into the "short" group, the essays between one hundred and three-hundred-fifty words into the "medium" group, and essays that were longer than three-hundred-fifty words into the "long" group. Then, we copied and pasted each essay into GPTZero to check for AI written work. After inputting the essays, GPTZero returned the percentage chance that the essays were written by generative AI. We wrote down the chance next to the word count of the essay inputted.

Lastly, we calculated the average percentage chance returned by GPTZero for each category of essays. In total there were six categories: short essays written by humans, medium essays written by humans, long essays written by humans, short essays written by AI, medium essays written by AI, and long essays written by AI.

## 4. Literature Review

There have been many studies on AI detectors. For example, a study by William H. Walters [1] evaluated the accuracy of sixteen AI detectors by inputting forty-two undergraduate essays written by ChatGPT-3.5, ChatGPT-4, and forty-two essays written by first year students into multiple AI detectors. Apart from other AI detectors, the paper found that GPTZero correctly identified eighty-one percent of all human and AI papers, while wrongfully identifying four percent [1]. Similar to Walters [1], we compared human vs. AI papers to examine false positives outputs by GPTZero. We used their survey as a precursor to shape the structure of our study.

Additionally, in a separate review conducted by Jae Liu and other colleagues [2], they analyzed fifty rehabilitation-related articles from four peer-reviewed journals, and fifty articles written using ChatGPT using various AI detectors. Out of all of the AI detectors tested, they found that GPTZero was able to correctly identify seventy percent of AI written papers as AI [2]. Similarly to us, they specifically used GPTZero, however, they included five additional papers. In a third survey conducted by Ahmed Elkhatat and his colleagues [3], fifteen paragraphs each from ChatGPT 3.5 and 4, and five human-written control responses were tested using various AI detectors. Their findings showed that AI detectors were more successful at identifying ChatGPT-3.5 as AI than ChatGPT-4 [3]. Although similar to our experiment, this study specifically examined the performance of AI detection tools across multiple ChatGPT models (ChatGPT-3.5 vs ChatGPT-4). It also presented us with adding extra paragraphs rather than papers in this experiment, so that we can expand on the word length.

Moreover, another survey analyzed over a hundred research articles to assess the accuracy of AI detectors using content from the medical field. The survey included behavioral health and psychiatry journals written between 2016-2018, and a hundred essays generated by Claude and ChatGPT. The study found that commercial AI detection software, such as originality.ai, was not as exceedingly accurate as people likely assumed (Popkov & Barrett, 2024) [4].

In addition, a research project that also looked into AI generated text detection, made use of and compared six different AI detectors (Akram, 2024) [5]. Similarly, another experiment looked into the capability of Turnitin AI detection tools. This experiment created twenty submissions using Chat-GPT and submitted them into Turnitin. Turnitin marked around ninety-five percent of the essays as containing AI, while only forty-five percent were marked as generated using AI (Perkins et al., 2023) [6]. This was an important experiment to source and notice, as it used Turnitin, an uncommonly known AI detector, providing diversity and a strong predecessor for our research. Lastly, another study looked into Machine-Obfuscated Plagiarism, or the act of altering texts generated from AI to make plagiarism detection harder, by comparing the effectiveness of six AI detectors (Foltýnek et. al., 1970) [7]. It was concluded that if a plagiarist uses OPT to reword a couple paragraphs, it causes the similarity to be below the Turnitin's threshold to be undetected.

All of these sourced studies gave us insight into how we can conduct our experiment. It helped us understand how we can determine the accuracy between page length and false positives, and by incorporating the methods utilized in each of these individual experiments, we were able to construct how many papers we needed for data, and how a correlation can be found between the entities we sought to find.

## 5. Results

Looking at Table 1 below, we found that GPTZero was more likely to wrongly identify human essays that were between zero and one-hundred words and human essays that were longer than three-hundred-fifty words. However, GPTZero was also proficient at detecting AI written work, regardless of length. For almost all lengths of essays written by AI, GPTZero was able to correctly identify the essays as written by AI. For example, for short essays, on average, GPTZero said that the given essays were 99% likely to be written by AI. For all AI categories, this chance never went below 98%.

Table 1 and Chart 1 below show the observed counts for GPTZero. Each cell represents the average percentage chance that an essay inputted was written by AI, according to GPTZero. For example, the cell that is intersected by "Human" and "Short Essays (0-100)" shows that, for all the essays inputted into GPTZero, on average, GPTZero thinks that there is a 29.86% chance that the essays were written by AI. In an ideal world, we would expect that chance for all human essays to be 0%, and we would expect that chance for all AI essays to be 100%. Table 2 and Chart 2 below show those chances. Overall, it seems that, as mentioned above, GPTZero is exceptional at correctly identifying essays written by AI, but is dreadful at correctly identifying essays written by humans.

*Table 1: GPTZero's assessment of the likelihood that a given essay was written by AI, according to our results.*

| % chance that a given essay was written by AI | Short Essays (0-100) | Medium (100-350) | Long (350-800) |
|---|---:|---:|---:|
| **Human** | 35.56 | 10.29 | 14.75 |
| **ChatGPT 3.5 & 4o** | 99.17 | 97.00 | 98.83 |

*Chart 1: GPTZero's assessment of the likelihood that a given essay was written by AI, according to our results, in graph form.*

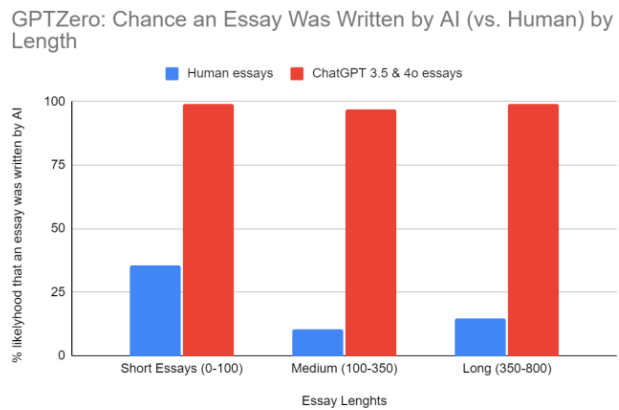

*Table 2: Expected counts in an ideal world for GPTZero's assessment of the likelihood that a given essay was written by AI.*

| % chance that a given essay was written by AI | Short Essays (0-100) | Medium (100-350) | Long (350-800) |
|---|---:|---:|---:|
| **Human** | 0 | 0 | 0 |
| **ChatGPT 3.5 & 4o** | 100 | 100 | 100 |

*Chart 2: Expected counts in an ideal world for GPTZero's assessment of the likelihood that a given essay was written by AI, in graph form.*

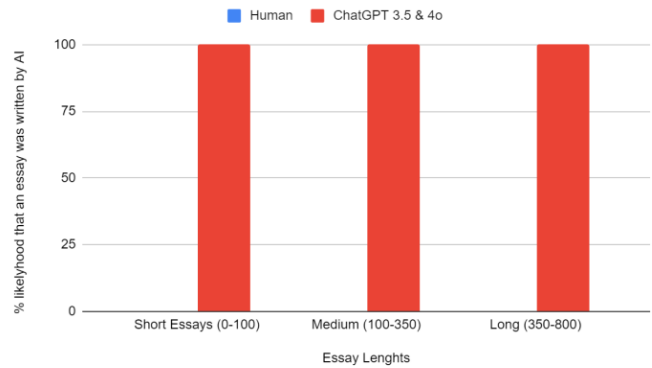

The confusion matrix in Table 3 illustrates GPTZero's performance in classifying AI vs. human written texts. It can be seen that although GPTZero was able to classify AI written essays correctly 100% of the time, it had some trouble with human essays. Of the fifty human-written essays, forty-two were accurately labeled as human, while eight were misclassified as AI. GPTZero's false positive rate was 16% (eight out of fifty human essays), contributing to an overall error rate of 10.3% (eight misclassifications out of seventy-eight total essays). Overall, GPTZero demonstrated excellent accuracy in detecting AI-generated content, but less consistency in correctly identifying human-written work.

*Table 3: Confusion Matrix.*

|  | **Predicted as Human** | **Predicted as AI** |
| --- | --- | --- |
| **Actually Human** | 42 | 8 |
| **Actually AI** | 0 | 28 |

## 6. Conclusion

Our study focused on the impact of certain essay lengths on the accuracy of GPTZero in detecting AI percentage. Using various essays from word counts between forty and eight-hundreds words, seventy-eight essays were submitted into GPTZero combined from human-generated and AI-generated essays. The purpose of our study was to test the accuracy of GPTZero's identification of AI-detected writing based on the lengths of given texts. Through our study, it was found that GPTZero can almost always correctly identify AI generated texts, at around 90-99% accuracy rate. However, human-generated essays, varying in lengths short, medium, and long, fluctuated significantly.

For example, short and long human-generated texts entered into GPTZero were found to be very inaccurate and/or had many false positives, while medium length texts were found to be identified the most accurately. However, regardless of the results, the constant fluctuations in the data indicate that medium length texts cannot be accurately predicted every time. More research would still need to be conducted to find a solid correlation and conclusion. The question still stands, is there any specific factor that causes false positives in AI generation, or is it merely a randomly generated percentage? For future research, it may be important to focus on generating even more texts, from both human and AI, including mixed texts as well (both human and AI), and focusing on other factors, such as page count instead of word count, or font size, etc.